\documentclass[conference]{IEEEtran}
\IEEEoverridecommandlockouts
\usepackage{cite}
\usepackage{amsmath,amssymb,amsfonts}
\usepackage{algorithmic}
\usepackage{graphicx}
\usepackage{textcomp}
\usepackage{xcolor}
\usepackage{filecontents}
\usepackage{hyperref}
\usepackage{multirow}
\usepackage{booktabs}
\usepackage{fullpage}
\usepackage{times}
\usepackage{array}
\usepackage{fancyhdr,graphicx,amsmath,amssymb}
\usepackage[ruled,vlined]{algorithm2e}
\include{pythonlisting}

\def\BibTeX{{\rm B\kern-.05em{\sc i\kern-.025em b}\kern-.08em
    T\kern-.1667em\lower.7ex\hbox{E}\kern-.125emX}}
\begin{document}

\title{TextDecepter: Hard Label Black Box Attack on Text Classification\\}

\author{\IEEEauthorblockN{Sachin Saxena}
\IEEEauthorblockA{\textit{Computer Science} \\
\textit{Rutgers University}\\
Camden, USA \\
ss3157@scarletmail.rutgers.edu}}

\maketitle

\begin{abstract}
Machine learning has been proven to be susceptible to carefully crafted samples, known as adversarial examples. The generation of these adversarial examples helps to make the models more robust and gives us an insight into the underlying decision-making of these models. Over the years, researchers have successfully attacked image classifiers in both, white and black-box settings. However, these methods are not directly applicable to texts as text data is discrete. In recent years, research on crafting adversarial examples against textual applications has been on the rise. In this paper, we present a novel approach for hard-label black-box attacks against Natural Language Processing (NLP) classifiers, where no model information is disclosed, and an attacker can only query the model to get a final decision of the classifier, without confidence scores of the classes involved. Such an attack scenario applies to real-world black-box models being used for security-sensitive applications such as sentiment analysis and toxic content detection.
\end{abstract}
\footnotetext{The code and links to the pretrained models are available at \href{https://github.com/SachJbp/TextDecepter}{https://github.com/SachJbp/TextDecepter} }

\section{Introduction}
Machine learning has shown superiority over humans for tasks like image recognition, speech recognition, security-critical applications like a bot, malware or spam detection. However. machine learning has been proven to be susceptible to carefully crafted adversarial examples. In recent years, research on the generation and development of defenses against such adversarial examples has been on the rise. These adversarial examples help to make the models more robust by highlighting the gap between sensory information processing in humans and the decisions made by machines. Attack algorithms have been formulated for image classification problems by \cite{carlini2017magnet}, \cite{Goodfellow2014}
. A classic example of an adversarial attack is that of a self-driving car crashing into another car because it ignores the stop sign. The stop sign being an adversarial example that an adversary intentionally replaced for the original stop sign. An example in the textual domain can be that of a spam detector that fails to detect a spam email. The spam email is an example of an adversarial attack in which the attacker has intentionally changed a few words or characters to deceive the spam detector. \par
The attacks can be broadly classified based on the amount of information available to the attacker as black-box or white-box attacks. White box attacks are those in which the attacker has full information about the model’s architecture, model weights, and the examples it has trained. Black-box attacks refer to those attacks in which only the final output of the model is accessible to the attacker. Black-box attacks can be further classified into 3 types based. The first type involves those attacks in which the probability scores to the outputs are accessible to the attacker referred to as the ’score-based black-box attacks’. The second type of attack involves the case where information of the training data is known to the attacker. The third attack type is the one in which only the final decision of the classifier is accessible by the attacker, with no access to the confidence scores of the various classes. We shall refer to the third type of attack as the Hard Label Black-Box attack.\par
In the NLP domain, researchers have mainly formulated attacks in the white box setting\cite{ebrahimi2017hotflip},\cite{li2018textbugger} with complete knowledge of gradients or the black-box setting with confidence scores accessible to the attacker\cite{gao2018black},\cite{li2018textbugger},\cite{alzantot2018generating},\cite{jin2019bert}. As per our knowledge, there has been no prior work done to formulate adversarial attacks against NLP classifiers in the hard label black-box setting. We emphasize hard label black-box attacks to be an important category of attacks much relevant to the real-world applications as the confidence scores can easily be hidden to avoid easy attacks.\par
Adversarial attacks on Natural Language models have to fulfill certain rules to qualify them as a successful attack: (1) Semantic Similarity- meaning of the crafted example should be the same as that of the original text, as judged by humans (2) Syntactic correctness: The crafted examples should be grammatically correct (3) Language Fluency: The generated example should look natural. \cite{jin2019bert} proposed TextFooler which aimed at preserving these three properties while generating adversarial examples. \par
We focus on the text classification task which is used for sentiment analysis, spam detection, topic modeling. Sentiment analysis is widely used in the online recommendation systems, where the reviews/comments are classified into a set of categories that are useful while ranking products or movies \cite{medhat2014sentiment}. Text classification is also used in applications critical for online safety like online toxic content detection \cite{nobata2016abusive}. Such applications involve classifying the comments or reviews into categories like irony, sarcasm, harassment, and abusive content. \par
Adversarial attacks on text classifiers consist of two main steps. First, identifying the important words in the text. Second, introducing perturbations in those words. For finding the important words, gradients are used in the white box setting and confidence scores in the black-box setting. In the absence of gradients or confidence scores in the hard-label black-box setting, it is non-trivial to locate important words. In the second step, there can be either character level perturbation, like, introducing space, replacements with visually similar characters, or word level perturbation by replacement of the word with its synonym. The selection of a character-level perturbation or synonym for replacement is based on the decrease in confidence value of the original class on introducing the perturbation. In the absence of confidence scores, there are no direct indicators to help select any type of perturbation unless the replacement of any single word leads to misclassification of the entire text, which is not always the case. \par
In our work, we come up with heuristics to help determine the important sentences and words in the first step and, select appropriate perturbation in the second step. The heuristics help us select a synonym for a replacement for each of the important words, in such a way that with each successive replacement we move towards the decision boundary.

Our main contributions are as follows:
\begin{enumerate}
\item We propose a novel approach to formulate natural adversarial examples against NLP classifiers in the hard label black-box setting.
\item We test our attack algorithm on three state-of-the-art classification models over two popular text classification tasks.
\item We improve upon the grammatical correctness of the generated adversarial examples.
\item We also decrease the memory requirement for the attack as compared to published attack systems involving word-level perturbations.
\end{enumerate}

\section{Attack Design}

\subsection{Problem Formulation}
Given an input text space $X$ and a set of $n$ labels , $Y = \{Y_1, Y_2, ..., Y_n\}$, we have a text classification model $F: X \rightarrow Y $ which maps from the input space $X$ to the set of labels $Y$. Let there be a text $x \in X$ which is correctly predicted by the model to be belonging to the class $y \in Y$ i.e. $F(x) = y$. We also have a semantic similarity function $Sim: X \times X \rightarrow [0,1] $. Then, a successful adversarial attack changes the text $x$ to $x_{adv}$, such that $$F(x_{adv}) \neq F(x)$$ $$ Sim (x, x_{adv}) \geq \epsilon$$ where $\epsilon$ is the minimum similarity between original and adversarial text.

Let us consider a binary classification model with labels $ \{Y_0, Y_1\}$ which we want to attack. We are given a text $T \in X$ having $m$ sentences, $S = \{s_1, s_2, ..., s_m\}$. The classification model correctly predicts $T$ to be belonging to class $Y_0$ i.e. $F(T) = Y_0$. We input each of the $m$ sentences to $F$ and get their individual labels. Let $A$ and $B$ form a partition of $S$ such that $F(a) = Y_0, \forall a \in A$ and $\mathcal{P}$ (A) be the power set of $A$. Let $\mathcal{A}_r$ be a set of $r$-combinations of $A$. We define another set $G = \{B \cup c \ | c \in \mathcal{P} (A) \} \}$. We refer to each of the elements in $G$ as an '\textbf{aggregate}'.

\subsection{Threat Model}

We consider the attack in the black-box setting, where an attacker does not have any information about the model weights or architecture and is allowed to query the model with specific inputs and get the final decision of the classifier model as an output. Further, the class confidence scores are not provided to the attacker in the output, making it a hard-label black-box attack. Although NLP APIs provided by Google, AWS and Azure do provide the confidence scores for the classes, but in a real-world application setting, like toxic content detection on a social media platform, the confidence scores are not provided, thereby making it a hard label black-box setting. Such an attack scenario also helps to gauge the model robustness.

\subsection{Methodology}

The proposed methodology for generating adversarial text has three main steps:

\textbf{Step 1: Sentence Importance Ranking}: We observe that when people convey opinions or emotions, not all the sentences convey the same emotion, few sentences are just facts without any emotion or sentiment. Other sentences can also be stratified based on varying levels of intensity. This forms the basis of our sentence ranking algorithm which helps to prioritize our attack on specific portions of the text in order of importance. \par
We assume that different sentences in the text contribute to the overall class decision to a varying level of intensity. Each of the sentences can either support or oppose the final decision of the classifier and the intensities which they do so are additive. Consider an example of sentiment analysis, where the labels are positive and negative.
The assumption of additivity of sentence class intensity, also helps us to infer that sentences in set $B$ when joined together to form a text, will belong to class $Y_1$. Hereby, we refer to the same as the class of set $B$ or the classifier's decision of set $B$. \par
We define the importance of a sentence in set $A$ by its ability to change the classifier's decision of set $B$ from $Y_1$ to $Y_0$. If an individual sentence from set $A$ when added to set $B$ is able to change the class of set $B$ from $Y_1$ to $Y_0$, then we consider the sentence to belong to \textbf{\textit{level 1 importance}}. More generally, if a sentence from set $A$ is able to change the classifier's decision of set $B$ only when it is put together with some subset of $A$ with at least $k-1$ sentences, then the sentence belongs to \textit{\textbf{level k importance}}. The $k-1$ other sentences in all such subsets also belong to \textit{\textbf{level k importance}}. Also, once the importance of a sentence is fixed at the $k^{th}$ level we do not consider it in subsequent levels. \\

{\SetAlgoNoLine%
\begin{algorithm}
\KwIn{Original Sentences set $S$ , ground truth label $y_0$, classifier $F(.)$ }
\KwOut{Sentence Importance Ranking}
 SentsSentiment $\leftarrow$ Find original labels of sentences in $S$ \;
 OrigLabelSents $\leftarrow$ Sentences with Label $Y_0$ \;
 OtherLabelSents $\leftarrow$ Sentences with Label $\neq Y_0$ \;
Let OrigLabelSents$_P$ represent set of all $P$ sentence combinations from OrigLabelSent \;
OrigLabelAggregates $\leftarrow \phi$ \;
 $P \leftarrow 1$ \;
 \While{OrigLabelSents}
 { TopSentImp$_P$ $\leftarrow \phi$ \;
 
 \For{Comb in OrigLabelSents$_P$}
 {AGG $\leftarrow$ Add $Comb$ to $OtherLabelSents$ and join it to form string  \;
 \If{$F(AGG)  = Y_0$}
 { OrigLabelAggregates $\leftarrow$ Add $AGG$\;
 \For{sent in Comb} 
{ SentImp[$sent$] $\leftarrow$ P \;

 TopSentImp$_P$ $\leftarrow$ Add $sent$}} 
 }
 Delete sentences in TopSentImp$_P$ from OrigLabelSents \;

$P \leftarrow P + 1$ \;
\If{$P > Length(OrigLabelSents)$} 
{Add remaining sentences in OrigLabelSents to TopSentImp$_P$ \;}
}
\Return SentImp , OrigLabelAggregates
\caption{Sentence Importance Ranking}
\end{algorithm}}%

{\SetAlgoNoLine%
\begin{algorithm}

\KwIn{Original text $X$ , ground truth label $Y_0$, classifier $F(.)$, semantic similarity threshold $\epsilon$, cosine similarity matrix }
\KwOut{Adversarial example $X_{adv}$ }
 Initialization: $X_{adv} \leftarrow X$\;
 Segment $X$ into sentences to get set $S$\; 
 SentImp, OrigLabelAggs = GetSentenceImp(S)\;
 WordImpScores = GetWordImp (SentImp)\;
 Create a set $W$ of all words $w_i \in X$ sorted by the descending
 order of their wordImpScores\;
MISCLASSIFIED $\leftarrow$ False \;
 \For{$w_j$ in $W $} {
\If{MISCLASSIFIED}{ break\;}
 CANDIDATES = GetSynonyms($w_j$)\;
 CANDIDATES $\leftarrow$ POSFilter (CANDIDATES) \;
 CANDIDATES $\leftarrow$ SEMANTICSIMFilter (CANDIDATES) \;
 FINCANDIDATES $\leftarrow$ Sort CANDIDATES by semantic similarity \;
 CHANGED $\leftarrow$ False  \;
 \For{$c_k$ in FINCANDIDATES}{
 $X' \leftarrow $ Replace $w_j$ with $c_k$ in $X_{adv}$ \;
 
 \If{$F(X') \neq Y_0$}{
   $X_{adv} = X'$ \;
   CHANGED $\leftarrow$ True \;
 MISCLASSIFIED $\leftarrow$ True \;
   }
 \If{NOT CHANGED}
 { SENT $\leftarrow$ Get the sentence in which $w_j$ belongs s.t  $F$(SENT) = $Y_0$ \;

 $X' \leftarrow $ Replace $w_j$ with $c_k$ in SENT\; 
 \If{$F(X')\neq Y_0$ \;}
 {$X_{adv} = X'$\;
  CHANGED $\leftarrow$ True  \;}}
 \If{NOT CHANGED}
 {OrigLabelAGG $\leftarrow$ Get the aggregate in which $w_j$ belongs s.t.  $F(AGG) = Y_0$ \;

 $X' \leftarrow $ Replace $w_j$ with $c_k$ in OrigLabelAGG \; 
 \If{$F(X')\neq Y_0$ \;}
 {$X_{adv} = X'$\;
 CHANGED $\leftarrow$ True \; 
 } 
 }
 }}
\caption{TextDecepter}
\end{algorithm}}%

\textbf{Step 2: Word importance ranking}:

After finding the importance of sentences in step 1, we need to find the importance ranking of the words to be attacked in these sentences. We observe that words with a certain Part of Speech (POS) tags are more important than others. For example, for a sentiment classification task, adjectives, verbs, adverbs are more important than nouns, pronouns, conjunctions, or prepositions. Further, we consider adjectives to be more important than adverbs. Consider a sentence, “The movie was very bad”. In this sentence, “bad” is the adjective and shapes the sentiment of the sentence. The adverb “very” increases the intensity of the adjective, making the predicted class confidence score increase further. \\

\textbf{Step 3: Attack}: We use the word Level perturbations in order of word importance obtained from the previous step. We select synonyms to replace the original words using cosine distance between word vectors. Further, to maintain the syntax of the language, only those synonyms having the same POS tags as that of the original word are considered for further evaluations. Experiments are done using, both, coarse and fine POS tag masks. \par

Let us look at the details of each of these steps:

\begin{enumerate}
\item Synonym extraction: We use the word embedding by \cite{mrkvsic2016counter} which obtained state-of-the-art performance on SimLex-999, a dataset designed by \cite{hill2015simlex} to measure how well different models judge the semantic similarity between words.
\item POS checking: To maintain the syntax of the adversarial example generated, we filter out the synonyms which have a different POS tag than the original word. We experiment with, both, coarse and fine POS tagging.
\end{enumerate}
We select a synonym to replace a particular word if its usage leads to one of the following:

\begin{enumerate}

\item Misclassification of the entire text.
\item Misclassification of the original labeled sentence to which the target word belongs.
\item Misclassification of the original label aggregate to which the target word belongs.

\end{enumerate}

\par If multiple synonyms fulfill the rules, then the one which fulfills the rule of higher preference is selected. If multiple synonyms fulfill the highest preference rule, then that synonym is selected whose placement in the review is semantically nearest to the original review.
We terminate the algorithm once the text misclassifies, or when all the important words have been iterated over.\par

The justification for the higher preference of sentence than the aggregate to which it belongs comes from the additivity assumption. Consider a sentence $v \in A$, $\{v\} \in \mathcal{A}_r$. Now, add it to set $B$ to form an 'aggregate', i.e. $B \cup \{v\} \in G$. Assuming the additivity of class intensities of sentences, we can easily see that when sentences in $B \cup \{v\}$ are joined to form a piece of text, it either belongs to class $Y_1$ or, in case, it belongs to class $Y_0$, then the intensity of class $Y_0$ is lesser when compared to $v$ alone. In other words, an 'aggregate' belonging to class $Y_0$ has lesser intensity of class $Y_0$ when compared with individual sentences belonging to class $Y_0$ which are part of that aggregate. Hence, a synonym which can flip the decision of the classifier, both, on the sentence and the aggregate (initially classified as $Y_0$) to which the sentence belongs would be preferred against a synonym which is just able to change the classifier's decision on the aggregate alone. \\ \par

\textbf{Step 4: Reset insignificant replacements} \\
Consider a directed acyclic graph, where a node represents a text and a directed edge between two nodes means that the latter node text can be obtained by replacement of a word $w$ in former node text by its most semantically similar synonym $w'$, given that the synonym has the same POS tag as the original word. Further, the weight of each edge is 1. Then, crafting an adversarial example from a given piece of text can be formulated as a path finding problem as follows; find a path from an initial node $T$ to final node $T'$, such that $F(T) = Y_0$ and $F(T') \neq Y_0$. The sentence and word ranking algorithm decide the order of perturbation. Thereafter, we use the aggregates belonging to the original class as heuristics to guide our graph search. The algorithm stops once the text misclassifies. Although, the heuristics help us in finding a path, but they do not guarantee if it is the shortest path. In the absence of confidence scores of classes, the problem of finding the shortest path has an exponential search space. So, once the given text is turned adversarial, the algorithm visits all the replaced words and resets those replacements without which the perturbed text remains adversarial.

\subsection{Time Complexity}
Let there be $n$ sentences in a text and $m$ words in total. 
\begin{enumerate}
\item Getting the class label of each of the individual sentences: $O(n)$
\item Sentence Importance ranking: The algorithm takes all possible combinations of sentences from the set of original labeled sentences and insert each of them to the set of sentences which do not have the original label. Hence, the time complexity of this step becomes $O(2^n)$. 
\item Word importance ranking:
\begin{enumerate}
\item Sort the original labeled sentences in order of their importance ranking: $O (n \log n)$
\item Iterate over all the words in the sorted sentences and make the word perturb sequence: $O(m)$
\item Generation of cosine similarity matrix of words in text with 65 thousand words in the vocabulary:  $O(m)$
\end{enumerate}
\item Pick the top $k$ synonyms for each of the candidate words that are to be perturbed. The cosine similarity matrix consists of similarity of each of the words in the text with all the words in the vocabulary. We sort the similarity values for each word with all the other words in the vocabulary. Hence, amounting to total time complexity of $O(m)$ 
\item Attack: Replace each of the $k$ synonyms of a word in the original text, the original labeled aggregates (obtained from sentence importance ranking step), and the original labeled sentences to which they belong.  Hence, the time complexity of this step becomes $O(kmn)$.  
\item Removal of insignificant perturbations: $O(m)$
\end{enumerate}

\par Hence, the overall time complexity of the attack framework becomes $O(2^n)+ O(kmn)$

\subsection{Memory requirement}
The generation of adversarial texts requires that we are able to get the synonyms for a particular word during the attack phase. Hence, a cosine similarity matrix is kept in the RAM to quickly generate synonyms for a particular word. In our attack framework, we keep the cosine similarity of only the words in the text that is being attacked, with all the 65 thousand words in the vocabulary. It is an improvement over the memory requirement in the attack framework of TextFooler \cite{jin2019bert} which keeps the cosine similarity matrix of all the 65 thousand words in the vocabulary, constituting a square matrix of size 65 thousand, consuming nearly 17 GB of RAM if single-precision floating-point format is used.

\section{Attack Evaluation: Sentiment Analysis}
We evaluate our attack methodology on generating adversarial texts for sentiment analysis tasks. Sentiment Analysis is a text classification task that identifies and characterizes the sentiment of a given text. It is widely used by businesses to get the sentiment of customers towards their product or services, by analyzing reviews or survey responses.

\subsection{Datasets and Models}
We study the effectiveness of our attack methodology on sentiment classification on IMDB and Movie Review (MR) datasets. We target three models: word-based convolutional neural network (WordCNN) \cite{kim2014convolutional}, word-based long-short term memory (WordLSTM) \cite{hochreiter1997long}, and Bidirectional Encoder Representations from Transformers (BERT) \cite{devlin2018bert}. We attack the pre-trained models open-sourced by \cite{jin2019bert} and evaluate our attack algorithm on the same set of 1000 examples that the authors had used in their work. We also run the attack algorithm against Google Cloud NLP API. The summary of the datasets used by \cite{jin2019bert} for training the models are in Table ~\ref{tab:table1} and their original accuracy are given in Table \ref{tab:table2}

\begin{table}[h]
\centering
\begin{tabular}{c|c|ccc} 
\hline
\textbf{Task}                            & \textbf{Dataset} & \textbf{Train} & \textbf{Test~} & \textbf{Avg Len}  \\ 
\hline
\multirow{2}{*}{\textbf{Classification}} & MR               & 9K             & 1K             & 20                \\
                                         & IMDB             & 25K            & 25K            & 215               \\
\hline  \\
\end{tabular}
\caption{Overview of the datasets used by \cite{jin2019bert} for training the models }
    \label{tab:table1}
\end{table}

\begin{table}[h]
\centering
\begin{tabular}{cccc} 
\hline
              & \textbf{wordCNN} & \textbf{wordLSTM} & \textbf{BERT}  \\ 
\hline
\textbf{MR}   & 79.9             & 82.2              & 85.8           \\
\textbf{IMDB} & 89.7             & 91.2              & 92.2           \\
\hline \\
\end{tabular}
\caption{Original accuracy of the target models on standard test sets}
    \label{tab:table2}
\end{table}

\subsection{Evaluation Metrics}

\begin{enumerate}

\item Attack success rate: We first measure the accuracy of the target model on the 1000 test samples and call it original accuracy. Then, we measure the accuracy of the target models against the adversarial samples generated from the same test samples and call it after-attack accuracy. The difference between the original accuracy and the after-attack accuracy of a classification model is called the attack success rate.
\item Percentage of Perturbed words: The percentage of words replaced by their synonyms on an average gives us a metric to quantify the change made to a given text.
\item Semantic similarity: It tells us the degree to which the two given texts carry a similar meaning. We use the Universal Sentence Encoder to measure the Semantic Similarity between original and adversarial text. Since my main aim is to generate adversarial texts, we just control the semantic similarity to be above a certain threshold.
\item Number of queries: The average number of queries made to the target model tells us the efficiency of the attack model.

\end{enumerate}

\section{Results}
\subsection{Automatic Evaluation}
We report our results of the hard label black-box attacks in terms of automatic evaluation on two text classification tasks using coarse and fine POS masks. The main results are summarized in Tables ~\ref{tab:table3} and \ref{tab:table11}. Our attack algorithm is able to bring down the accuracy of all the major text classification models with an attack success rate greater than 50\% for all the models. Further, the percentage of perturbed words is nearly 3\% for all the models on the IMDB dataset and between 10-16\% for all the models on the MR dataset. In the IMDB dataset, which has an average word length of 215 words, our attack model is able to conduct successful attacks by perturbing less than 7 words on average. That means that our attack model can identify the important words in the text and makes subtle manipulations to mislead the classifiers. Overall, our algorithm can attack text classification models pertaining to sentiment analysis with an attack success rate greater than 50\%, no matter how long the text sequence or how accurate the target model. Further, our model requires the least amount of information among all the models it is compared with. \par

The attack model is also able to attack GCP NLP API and brings down the accuracy from 76.4\% to 16.7\% for the MR dataset. Further, it changes only 10.2\% of the words in the text to generate the adversary. The results are unprecedented as we have achieved the results without having any information about the confidence scores of the classes involved. The attack algorithm and the carefully crafted adversarial texts can be utilized for the study of interpretability of the BERT model \cite{feng2018pathologies}. \par

The query number is almost linear to the text length, with a ratio in (6,10) which is at par with \cite{jin2019bert} and \cite{li2018textbugger}.

\begin{table*}[h]
\centering
\begin{tabular}{lcc|cc|cc|c} 
\hline
~                              & \multicolumn{2}{c|}{\textbf{wordCNN}} & \multicolumn{2}{c|}{\textbf{wordLSTM}} & \multicolumn{2}{c|}{\textbf{BERT}} & \textbf{GCP NLP API}  \\ 
\hline
                               & \textbf{MR} & \textbf{IMDB}           & \textbf{MR} & \textbf{IMDB}            & \textbf{MR} & \textbf{IMDB}        & \textbf{MR}           \\
\textbf{Original Accuracy}     & 78          & 89.4                    & 80.7        & 90.3                     & 90.4        & 88.3                 & 76.4                  \\
\textbf{After-attack accuracy} & 18.9        & 17.3                    & 18.9        & 32.5                     & 42.3        & 30.9                 & 16.6                  \\
\textbf{Attack Success rate}   & 75.8        & 80.6                    & 76.6        & 64.0                     & 53.2        & 65.0                 & 78.3                  \\
\textbf{\% Perturbed Words}    & 12.1        & 3.1                     & 12.2        & 2.8                      & 15.6        & 2.1                  & 11.8                  \\
\textbf{Query number}          & 133.2       & 1368.6                  & 123.2       & 1918.1                   & 189.5       & 1719.7               & 126.8                 \\ 
\hline
\textbf{Average Text Length}   & 20          & 215                     & 20          & 215                      & 20          & 215                  & 20                    \\
\hline \\
\end{tabular}

\caption{Automatic evaluation results on text classification datasets (using coarse POS mask)}
    \label{tab:table3}
\end{table*}

\begin{table*}[h]
\centering
\begin{tabular}{lcc|cc|cc|c} 
\hline
~                              & \multicolumn{2}{c|}{\textbf{wordCNN}} & \multicolumn{2}{c|}{\textbf{wordLSTM}} & \multicolumn{2}{c|}{\textbf{BERT}} & \textbf{GCP NLP API}  \\
                               & \textbf{MR} & \textbf{IMDB}           & \textbf{MR} & \textbf{IMDB}            & \textbf{MR} & \textbf{IMDB}        & \textbf{MR}           \\ 
\cline{2-8}
\textbf{Original Accuracy}     & 78          & 89.4                    & 80.7        & 90.3                     & 90.4        & 88.3                  & 76.4                  \\
\textbf{After-attack accuracy} & 20.7        & 18.9                    & 21.2        & 34.4                     & 45.9        & 33.3                    & 16.6                  \\
\textbf{Attack Success rate}   & 73.5        & 78.9                    & 73.7        & 61.9                     & 49.2        &  62.3          & 78.3                  \\
\textbf{\% Perturbed Words}    & 12.2        & 3.1                     & 12.0        & 2.6                      & 14.6        & 2.2                   & 10.2                  \\
\textbf{Query number}          & 112.5       & 1230.0                  & 107.0       & 1650.5                   & 159.0       & 1507.4                    & 109.6                 \\ 
\hline
\textbf{Average Text Length}   & 20          & 215                     & 20          & 215                      & 20          & 215                  & 20                    \\
\hline \\
\end{tabular}
\caption{Automatic evaluation results on text classification datasets (using fine POS mask)}
    \label{tab:table11}
\end{table*}

\subsection{Benchmark Comparison}
We compare our attack against state-of-the-art adversarial attack systems on the same target model and dataset. For GCP NLP API, we compare our attack results against \cite{li2018textbugger} and \cite{gao2018black} on MR datasets.
With wordCNN and wordLSTM as the target models, the comparison is against \cite{li2018textbugger}, \cite{alzantot2018generating}, \cite{jin2019bert}. The results of the comparison are summarised in table ~\ref{tab:table4} and ~\ref{tab:table5}. The lower attack success rates, when compared to the other attack systems, can be attributed to the fact that our attack system does not make use of confidence scores of the classes that other published systems do.

\begin{table*}[h]
\centering
\begin{tabular}{lcc} 
\hline
\textbf{Attack System}    & \textbf{Attack Success Rate} & \textbf{\%Perturbed Words}  \\ 
\hline
\textbf{Li et al \cite{li2018textbugger}}      & 86.7                         & 6.9                         \\ 
\hline
\textbf{Alzantot al \cite{alzantot2018generating}} & 97.0                         & 14.7                        \\ 
\hline
\textbf{Jin et al. \cite{devlin2018bert}}    & 99.7                         & 10.0                        \\ 
\hline
\textbf{Ours}             & \textbf{64.0}                & \textbf{2.4}                \\
\hline \\
\end{tabular}
\caption{Comparison of our attack system against other published systems with wordLSTM  as the target model (Dataset: IMDB) }
    \label{tab:table4}
\end{table*}
 
\begin{table*}[h]
\centering
\begin{tabular}{lccc} 
\hline
\multicolumn{1}{c}{\textbf{Attack System}} & \textbf{Original Accuracy} & \textbf{Attack Success Rate} & \textbf{\%Perturbed Words}  \\ 
\hline
\textbf{Gao et al. \cite{gao2018black}}                       & 76.7                       & 67.3                         & 10                          \\
\textbf{Li et al. \cite{li2018textbugger}}                        & 76.7                       & 86.9                         & 3.8                         \\
\textbf{Ours}                              & 76.4                       & 78.1                         & 10.2                        \\
\hline \\
\end{tabular} 
\caption{Comparison of our attack system against other published systems with Google Cloud NLP API  as the target model (Dataset: MR) }
    \label{tab:table5}
\end{table*}

\begin{table*}[h!]
\centering
\begin{tabular}{lll}
\hline
&                                & \textbf{Movie Review (Positive (POS) $\leftrightarrow$ Negative (NEG))}                                                                \\ 
\hline
\multirow{6}{*} & \textbf{Original (Label: NEG)} & i firmly \textit{\textbf{believe}} that a good video game movie is going to show up soon i also believe that resident evil is not it                                 \\
                                   & \textbf{Attack (Label: POS}    & i firmly \textit{\textbf{feel}} that a good video game movie is going to show up soon i also believe that resident evil is not it                                                                      \\ 
\cline{2-3}
                                   & \textbf{Original (Label: POS)} & strange and \textit{\textbf{beautiful}} film           \\
                                   & \textbf{Attack (Label: NEG)}   & strange and \textit{\textbf{resplendent}} film  \\ 
\cline{2-3}
                                   & \textbf{Original (Label: POS)} & the lion king was a roaring \textit{\textbf{success}} when it was released eight years ago , but on imax it \textit{\textbf{seems}} better, not just bigger                     \\
                                   & \textbf{Attack (Label: NEG}    & the lion king was a roaring \textit{\textbf{attainment}} when it was released eight years ago , but on imax it \textit{\textbf{transpires}} better , not just bigger                    \\
\hline \\
\end{tabular} 

\caption{Examples of original and adversarial sentences from MR (GCP NLP API) }
    \label{tab:table7}
\end{table*}

\subsection{Human Evaluation}
Following the practice of \cite{jin2019bert}, we perform human evaluation by sampling 100 adversarial examples from the MR dataset with the WordLSTM. We perform three experiments to verify the quality of our adversarial examples. First, the human judges are asked to give the grammaticality score of a shuffled mix of original and adversarial text on a scale of 1-5. As shown in Table, the grammaticality of the adversarial texts with fine POS tag mask is closer to the original texts when compared with coarser POS tag mask. However both of the scores are above 4 meaning that using, both, coarse and fine POS tags result in smooth adversarial texts. \par

Second, the judges assign classification labels to a shuffled set of original and adversarial texts, for both coarse and fine POS masks. The results show that the overall agreement between the labels of the original and adversarial text for both the cases are quite high, 92\% and 93\% respectively. This suggests that improving the grammaticality of the adversarial texts using a fine POS mask does not contribute much to the overall meaning of the texts to humans. \par

Third, the judges determine whether the adversarial texts retain the meaning of the original text. The judges are given three options, 1 for similar, 0.5 for ambiguous, and 0 for dissimilar. The average sentence similarity score is 0.88 when a fine POS mask is used compared to 0.86 when a coarse POS mask is used for synonym-selection, suggesting a marginal improvement in sentence similarity scores in the former.

\begin{table*}[h]
\centering
\begin{tabular}{lp{13cm}l} 
\hline
\
                               & \multicolumn{1}{c}{\textbf{ Movie Review (Positive (POS)$\leftrightarrow$ Negative (NEG))}}                                                                                                                                                                                                                                                                                                                                                                                                                                                                                                                                                                                                                                                                                                                                                                                                                                                                   \\ 
\hline
\\
\textbf{Original (Label: NEG)} & after the book i became very \textbf{\textit{sad}} when i was watching the movie . i am agree that sometimes a film should be different from the original novel but in this case it was more than acceptable . some examples: 1 ) why the ranks are different ( e.g. lt . diestl instead of sergeant etc.) 2 ) the final screen is very \textbf{\textit{poor }}and makes diestl as a soldier who feds up himself and wants to die . but it is not true in 100 \% . just read the book . he was a bull - dog in the last seconds as well . he did not want to die by wrecking his gun and walking simply towards to michael \& noah . so this is some kind of a happy end which does not fit at all for this movie .                                                                                                                                                                                                                                                     \\ \\
\textbf{Attack (Label: POS)}    & after the book i became very \textbf{\textit{bleak }}when i was watching the movie . i am agree that sometimes a film should be different from the original novel but in this case it was more than acceptable . some examples:1 ) why the ranks are different ( e.g. lt . diestl instead of sergeant etc.) 2 ) the final screen is very \textbf{\textit{flawed }}and makes diestl as a soldier who feds up himself and wants to die . but it is not true in 100 \% . just read the book . he was a bull - dog in the last seconds as well . he did not want to die by wrecking his gun and walking simply towards to michael \& noah . so this is some kind of a happy end which does not fit at all for this movie .~                                                                                                                                                                                       \\ 
\\
\hline 
\\
\textbf{Original (Label: NEG)}  & seriously , i do n't understand how justin long is becoming increasingly popular . he either has the best agent in hollywood , or recently sold his soul to satan . he is almost \textit{\textbf{unbearable}} to watch on screen , he has little to no charisma , and \textit{\textbf{terrible}} comedic timing . the only film that he has attempted to anchor that i 've remotely enjoyed was waiting ... and that is almost solely because i 've worked in a restaurant . but i digress . aside from it 's \textit{\textbf{terrible}} lead , this film has loads of other debits . i understand that it 's supposed to be a \textit{\textbf{cheap}} popcorn comedy , but that does n't mean that it has to completely insult our intelligence , and have writing so incredibly hackneyed that it borders on offensive . lewis black 's considerable talent is wasted here too , as he is at his most incendiary when he is unrestrained , which the pg-13 rating certainly wo n't allow . the film 's sole bright spot was jonah hill ( who will look almost unrecognizable to fans of the recent superbad due to the amount of weight he lost in the interim ) . his one liners were funny on occasion , but were certainly not enough to make this anywhere close to bearable . if you just want to completely turn your brain off ( or better yet , do n't have one ) then maybe you 'd enjoy this , but i ca n't recommend it at all .
\\ \\
\textbf{Attack (Label: POS)}   & seriously , i do n't understand how justin long is becoming increasingly popular . he either has the best agent in hollywood , or recently sold his soul to satan . he is almost \textit{\textbf{terrible}} to watch on screen , he has little to no charisma , and \textit{\textbf{spooky}} comedic timing . the only film that he has attempted to anchor that i ' ve remotely enjoyed was waiting ... and that is almost solely because i ' ve worked in a restaurant . but i digress . aside from it 's \textit{\textbf{spooky}} lead , this film has loads of other debits . i understand that it 's supposed to be a \textit{\textbf{miserly}} popcorn comedy , but that does n't mean that it has to completely insult our intelligence , and have writing so incredibly hackneyed that it borders on offensive . lewis black 's considerable talent is wasted here too , as he is at his most incendiary when he is unrestrained , which the pg-13 rating certainly wo n't allow . the film 's sole bright spot was jonah hill ( who will look almost unrecognizable to fans of the recent superbad due to the amount of weight he lost in the interim ) . his one liners were funny on occasion , but were certainly not enough to make this anywhere close to bearable . if you just want to completely turn your brain off ( or better yet , do n't have one ) then maybe you 'd enjoy this , but i ca n't recommend it at all .                                          \\
\hline
\\ \\
\end{tabular}

\caption{Examples of original and adversarial sentences from IMDB (BERT) }
    \label{tab:table8}
\end{table*}

\begin{table*}[h!]
\centering
\begin{tabular}{ccc} 
\hline
                     & \textbf{Fine POS filter} & \textbf{Coarse POS filter}  \\ 
\hline
\textbf{Original}    & 4.5                      & 4.5                         \\
\textbf{Adversarial} & 4.3                      & 4.1                         \\
\hline \\
\end{tabular}
\caption{Grammaticality of original and adversarial examples for MR (BERT) ON 1-5 scale}
    \label{tab:table9}
\end{table*}

\begin{table*}[h]
\centering
\begin{tabular}{ll} 
\hline
                                                   & \textbf{Movie Review (Positive (POS) $\leftrightarrow$ Negative (NEG))}                                 \\ 
\hline
\textbf{Original (Label: POS)}                     & she may not be real , but the \textbf{laughs }are                                                                \\
\textbf{Attack (Label: NEG) using coarse POS tags} & she may not be real , but the \textbf{\textit{kidding }}are                                                      \\
\textbf{Attack (Label: NEG) using fine POS tags} & she may not be real , but the \textbf{\textit{chuckles }}are                                                     \\ 
\hline
\textbf{Original (Label: NEG)}                     & falsehoods pile up , \textbf{\textit{undermining }}the movie 's reality and stifling its creator 's comic voice  \\
\textbf{Attack (Label: POS) using coarse POS tags} & falsehoods heaps up , \textbf{\textit{jeopardizes~}}the movie 's reality and stifle its creator 's comic voice   \\
\textbf{Attack (Label: POS) using fine POS tags} & falsehoods heaps up , \textbf{\textit{jeopardizing~}}the movie 's reality and stifle its creator 's comic voice  \\
\hline \\
\end{tabular} 

\caption{Qualitative comparison of adversarial attacks with coarse and fine POS tagging for synonyn selection. Target Model is wordLSTM }
    \label{tab:table6}
\end{table*}

\newcolumntype{C}[1]{>{\centering\let\newline\\\arraybackslash\hspace{0pt}}m{#1}}
\begin{table*}[h]
\centering
\begin{tabular}{cccccc}
              & \textbf{Orig Acc.} & \multicolumn{2}{C{2cm}}{\textbf{After-Attack accuracy}} & \multicolumn{2}{C{2cm}}{\textbf{\% Perturbed words}}  \\
              &                            & \textbf{w/ agg.} & \textbf{w/o agg.}     & \textbf{w/ agg.} & \textbf{w/o agg.}   \\ 
\hline
\textbf{IMDB} & 88.3                       & 30.9                & 63.2                          & 2.1                  & 0.6                       \\
\textbf{MR}   & 90.4                       & 42.4                 & 75                          & 13.6                 & 10.1                      \\
\hline \\
\end{tabular}
\caption{Comparison of the after-attack accuracies of the BERT model with and without using aggregates for synonym selection  }
    \label{tab:table10}
\end{table*}

\subsection{Comparison of Fine and Coarse POS tag filter}
Part of Speech (POS) tags can be of two types: Coarse-grained (Noun, Verb, Adjective, etc.) or Fine-grained (Noun-proper-singular, noun-proper-plural, verb-past, verb-present, etc). TextDecepter uses a POS mask to reject those synonyms which do not belong to the same POS as that of the original word when they are placed in the text. A Fine-grained POS mask helps to maintain grammaticality to a greater extent when compared to the coarse-grained POS mask as demonstrated in Table \ref{tab:table6}. However, during human evaluation, we observed that improving the grammaticality of the adversarial texts for sentiment classification dataset using fine POS does not contribute much to the overall understanding of the text for humans. We recommend using the appropriate POS mask depending on the grammaticality requirement for the context in which adversarial texts are being generated.

\section{Discussion}
\subsection{Ablation study}
\textbf{Aggregates}
The most critical step of our algorithm is the use of aggregates, which belong to the original class, to select or reject synonyms for replacement. To validate the effectiveness of this step we remove the usage of aggregates and select a synonym for replacement only when its presence is able to misclassify the original text. The results for the BERT model are shown in table \ref{tab:table10} . After removing the sentence importance ranking step, we see that the after-attack accuracy increases by 32\% for IMDB and 35\% for MR dataset, respectively. This suggests the importance of aggregates for selecting synonym for replacement, the removal of which renders the attack ineffective. The aggregates generated in the sentence importance ranking step help us to select those synonyms which can take the original text towards misclassification.

\section{Conclusion}
We propose a hard label black-box attack strategy for text classification tasks. We also conduct extensive experimentation on sentiment analysis datasets to validate our attack system. We also conduct human evaluation to validate the grammatical and semantic correctness of the generated adversarial examples. The attack algorithm uses the assumption of additivity of class intensities of sentences to craft adversarial examples and achieves attack success rate of more than 50 \% against state-of-the-art text classification models. The adversarial examples generated can be utilized to improve the existing text classification models for sentiment analysis by including them in the training dataset. The attack algorithm can also be used to gauge the robustness of text classification models pertaining to sentiment analysis. We also improved upon the memory requirement for the attack, as compared to the other comparative attack frameworks. As a future line of work, it would be imperative to run the attack algorithm on multi-class classification and textual entailment tasks to test the generalization capability of the proposed methodology.

\section*{Acknowledgment}
I thank Prof. Sunil Shende for insightful discussions. I especially appreciate Kalyan Alapati and Dheenadhyalan Kumaraswamy for helping with human evaluation.

\bibliographystyle{IEEEtran}
\bibliography{References}

\end{document}